\documentclass[runningheads]{llncs}
\usepackage{amsmath,amssymb}
\usepackage{color}
\usepackage[width=122mm,left=12mm,paperwidth=146mm,height=193mm,top=12mm,paperheight=217mm]{geometry}

\usepackage[table]{xcolor,pict2e}
\usepackage{placeins}
\usepackage[lofdepth,lotdepth]{subfig}

\usepackage[percent]{overpic}
\usepackage{stackengine}
\usepackage{graphicx}

\usepackage{hhline}
\usepackage[font=small,labelfont=bf,tableposition=top]{caption}

\usepackage{marvosym}

\definecolor{mygray}{gray}{0.6}

\newcommand{\etal}{\emph{et al.}}

\DeclareCaptionLabelFormat{andtable}{#1~#2  \&  \tablename~\ref{tab:res-kaggle}}

\begin{document}
\pagestyle{headings}
\mainmatter

\title{Multi-level Activation for Segmentation of Hierarchically-nested Classes}

\titlerunning{Multi-level Activation for Hierarchically-nested Classes}

\author{Marie~Piraud\inst{1,2} \and
Anjany~Sekuboyina\inst{1} \and
Bj\"orn~H.~Menze\inst{1}}
\institute{Department of Computer Science, Technische Universit\"at M\"unchen,\\
Munich, Germany,
\and
\email{marie.piraud@tum.de}}

\maketitle

\begin{abstract}
For many biological image segmentation tasks, including topological knowledge, such as the nesting of classes, can greatly improve results.
However, most `out-of-the-box' CNN models are still blind to such prior information.
In this paper, we propose a novel approach to encode this information, through a multi-level activation layer and three compatible losses.
We benchmark all of them on nuclei segmentation in bright-field microscopy cell images from the 2018 Data Science Bowl challenge,
 offering an exemplary segmentation task with cells and nested subcellular structures.
Our scheme greatly speeds up learning, and outperforms standard multi-class classification with soft-max activation and a previously proposed method stemming from it, improving the Dice score significantly (p-values~$<0.007$). 
Our approach is conceptually simple, easy to implement and can be integrated in any CNN architecture.
It can be generalized to a higher number of classes, with or without further relations of containment.
\end{abstract}

\section{Introduction}
For certain multi-class segmentation tasks, the classes have a hierarchical topological relation: one class is nested into another one, meaning that the set of pixels of the second class is spatially surrounded by pixels from the first one, as illustrated in Fig.~\ref{fig:act-2nested}(a).
This is in the case of several important biological and medical image analysis tasks: anatomical structures are organized along the anatomical tree, tumors are often contained in one particular organ or anatomical structure, or intracellular features follow a specific organization within the cell.
Informing the network about this type of structural relations between classes as a prior can significantly improve segmentation results, enabling the algorithm to focus on the hidden and unforseen features~\cite{Nosrati2014,BenTaieb2016}.
Convolutional Neural Networks (CNNs), which have become the state-of-the-art for most image segmentation applications, have proven to be able to learn and encode very complex structures and relations between objects. 
However, very few CNN models are able to encode topological information as a prior.

In the literature, most of it predating the widespread use of CNNs, we distinguish three main avenues that have been pursued with that objective: 
(i)~The first option is to use cascaded geometries~\cite{Christ2016}, by training independently successive segmentation networks, the first for the surrounding class, and the second for the nested one.
(ii)~A second option can be to modify the loss term to penalize predictions which do not respect the expected topology, either by modifying the cross-entropy loss taking into account label-relations~\cite{BenTaieb2016} or by integrating class relations through a specifically designed Wasserstein distance matrix in the Dice score loss~\cite{Fidon2017}; both methods relying on soft-max activation.
(iii)~A third option is integrating label context via Conditional~\cite{Bauer2014,Alberts2015}  and Markov~\cite{Liu2017} Random Fields that, although used as postprocessing routines in most application, can be integrated with deep learning architectures.
All aforementioned methods however handle the nesting of classes in a rather indirect way --~either in separate stages or through the loss, that often needs to be parametrized~-- and are therefore not optimally using information on class relations.
Moreover, soft-max activation and cross-entropy loss assume that the classes are mutually-exclusive, as a pixel cannot belong to several classes at the same time, which does not make a natural basis for classes with hierarchical topological constraints.
Applying such a standard method to segment nested-classes can lead to unreasonable results, with e.g. tumors detected outside of the organ of interest~\cite{Christ2016}, or nuclei at the border of the cell [see Fig.~\ref{fig:res-nuclei}], thereby limiting the quality of the results.

As a paradigm shift, 
we propose to consider the segmentation of hierarchically-nested classes as a generalized logistic regression problem by using a
 multi-level activation layer. 
This naturally and directly enforces the nesting of the classes, trading off neighbourhood constraints with local observations automatically and permits to segment all nested classes with a single output channel.
This novel activation requires to move away from traditional cross-entropy loss, such that we introduce three adapted loss functions, and show that they all greatly speed up the learning process, and perform better than standard multi-class classification and the method from Ref.~\cite{BenTaieb2016}, on nuclei segmentation in bright-field microscopy images.
We provide a second benchmark of our method on liver lesion segmentation in Computer Tomography (CT) images in the Supplementary Material. 
On top of being conceptually simple, the multi-level activation method is easy to implement, does not need parametrization and can be integrated in any CNN architecture.

\section{Method}
We start by describing our methodological contributions. We first introduce the new activation layer, and a matching thresholding scheme to infer the output segmentation map. We then propose three loss functions adapted to this activation, in Sec.~\ref{sec:losses}.

\subsection{Multi-level activation layer}
\begin{figure}[tb]
\raisebox{0.45cm}{\begin{overpic}[height=0.24\textwidth]{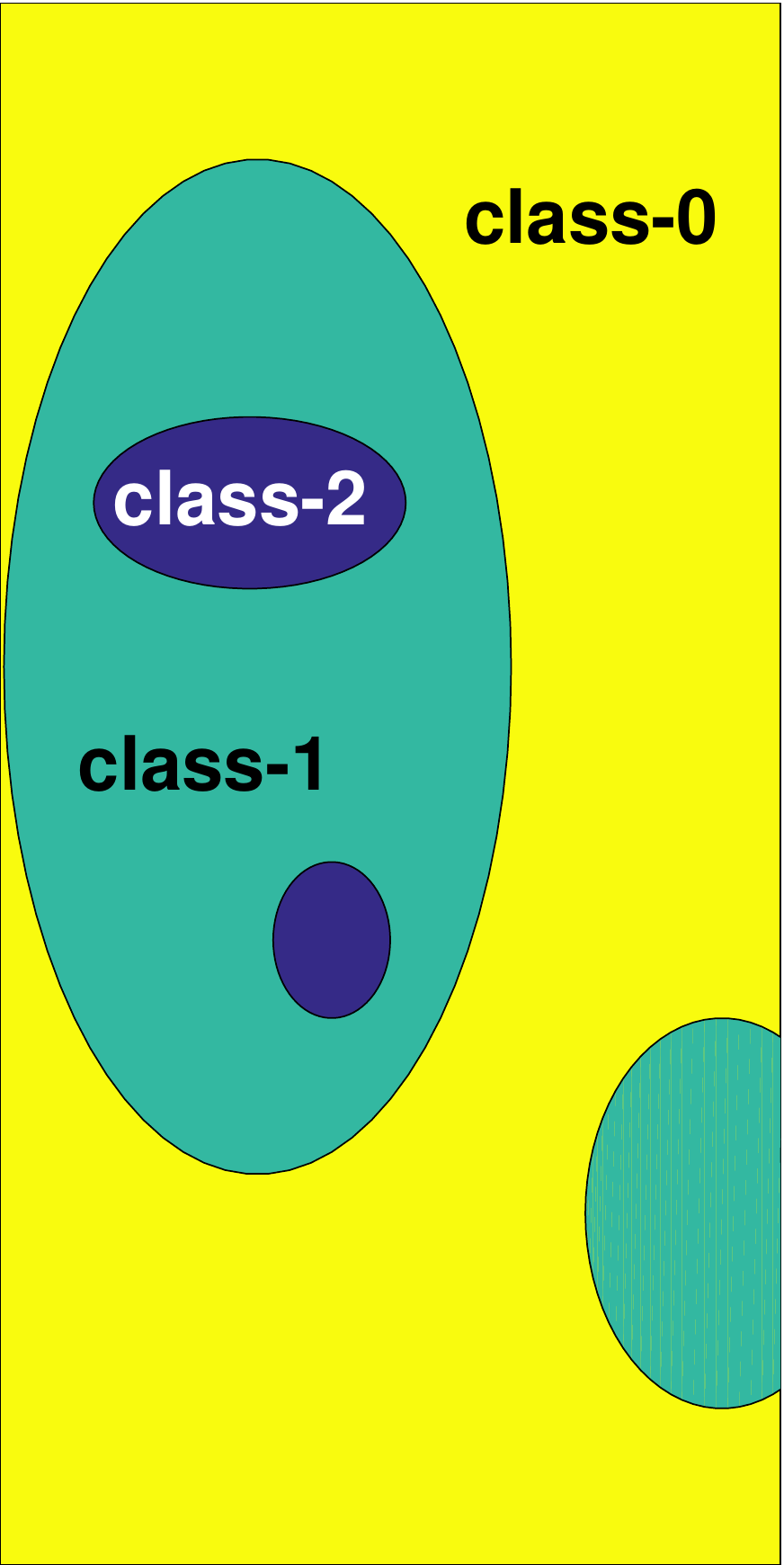}
 \put (2,90) {a)} \end{overpic}}
\hfill
\begin{overpic}[height=0.28\textwidth]{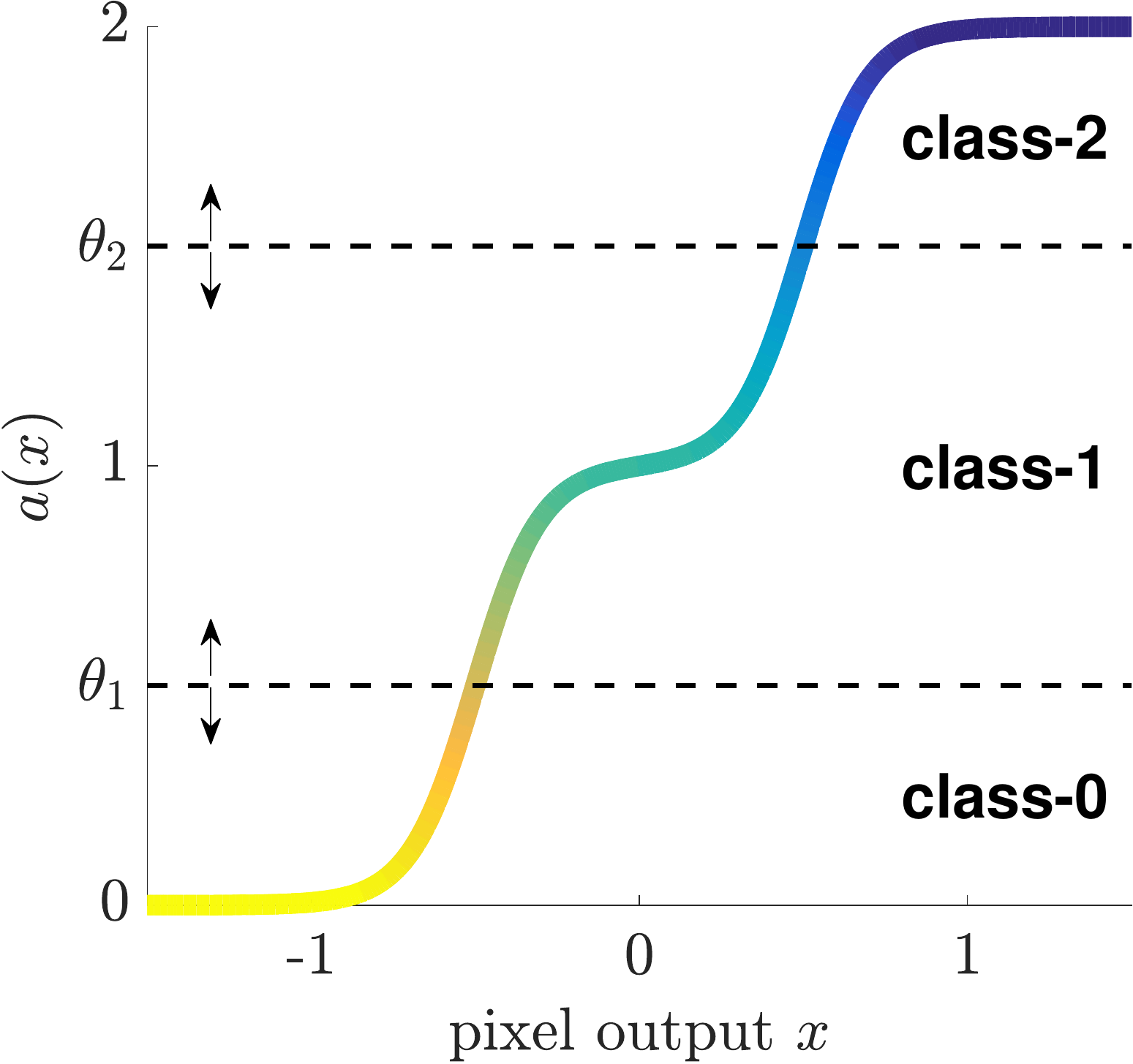}
 \put (-5,84) {b)} \end{overpic}
\hfill
\begin{overpic}[height=0.28\textwidth]{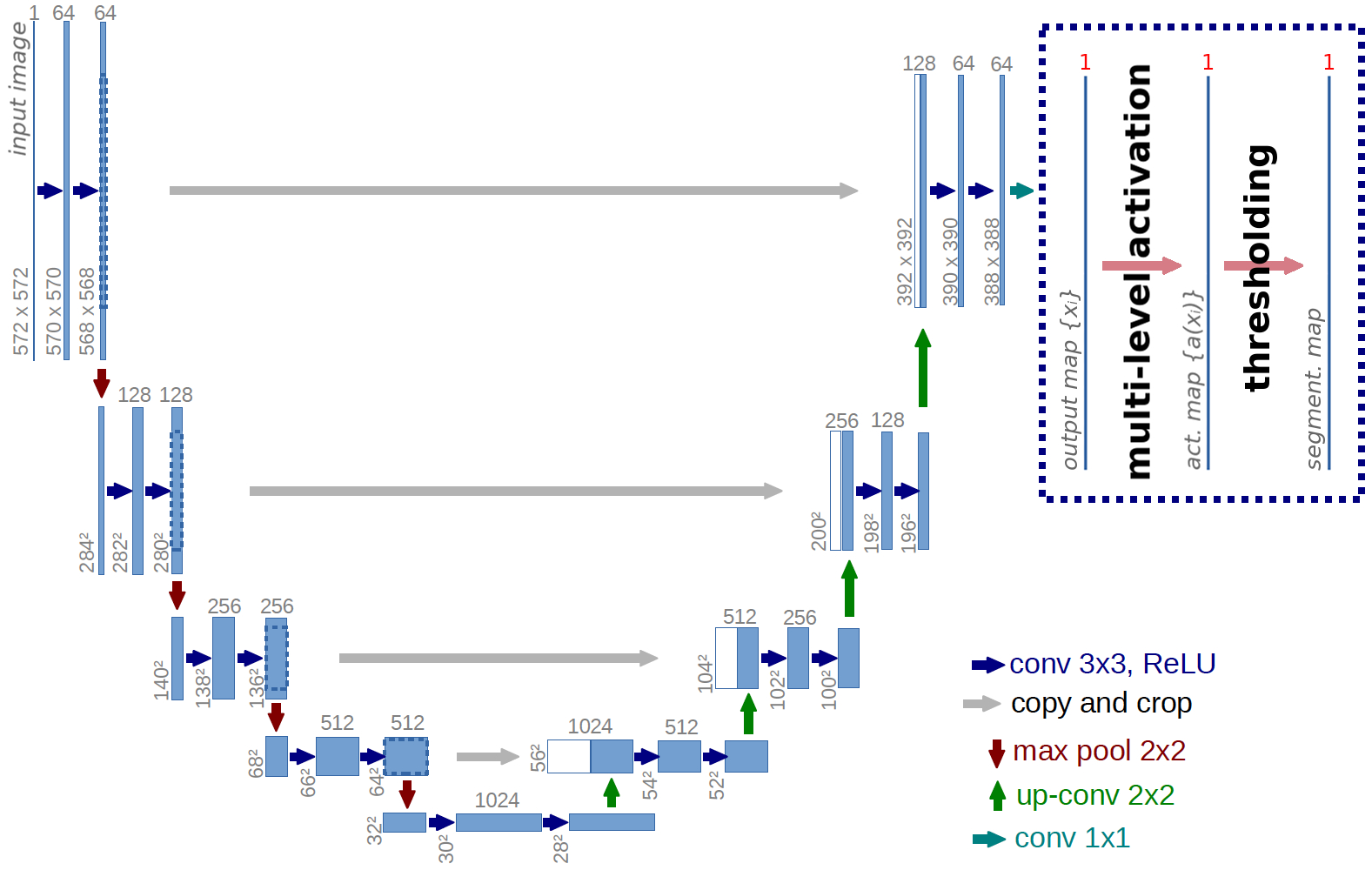}
 \put (-5,57) {c)} \end{overpic}
\caption{\emph{Illustration of the method}. (a) Sketch of 3 nested classes. (b) Corresponding multi-level activation [Eq.~\eqref{eq:act-2nested}, with $h=1$ and $\kappa=10$]. (c) Multi-level activation block, which can be implemented on top of any segmentation architecture, here the U-Net~\cite{Ronneberger2015}.}
\label{fig:act-2nested}
\end{figure}
Inspired by continuous regression, we propose a new multi-level activation layer, thereby generalizing logistic regression to hierarchically-nested classes [class-$m$ $\subset$ class-$(m-1)$ $\subset$ ... $\subset$ class-1 $\subset$ class-0]. 
This activation function should have the same number of levels as the number of classes $m+1$, we therefore construct it from $m$ equally-spaced sigmoids
\begin{equation}
a(x) = \sum_{n=1}^{m} \sigma \left( \kappa \left[ x+h \left( n-\frac{m+1}{2} \right) \right] \right)\, ,
\label{eq:act-2nested}
\end{equation}
where $\sigma$ is the sigmoid function, $\kappa$ its steepness and $h$ the spacing between consecutive sigmoids. 
A similar activation function has also been introduced for unsupervised RGB image segmentation~\cite{Bhattacharyya2007}.
In the case of $m+1=3$ classes, which is illustrated in Fig.~\ref{fig:act-2nested}(a), it becomes a two-level sigmoid 
$a(x) = \sigma \left[ \kappa (x+h/2) \right] + \sigma \left[ \kappa (x-h/2) \right]$.
This is illustrated in Fig.~\ref{fig:act-2nested}(b) for $h=1$ and $\kappa=10$, that we use in the following.
This pixel-wise activation layer is designed to replace the soft-max layer of any CNN architecture, see Fig.~\ref{fig:act-2nested}(c), enabling the segmentation of nested classes with one output channel, inherently respecting their hierarchy.
Note that it does not enforce the topology as a strict constraint in the segmentation map, but the hierarchy holds as long as the output map of the network remains smooth, which is the case if the resolution of the image is high enough.

Further generalizing logistic regression, we infer the output segmentation map from the activation map $a(x_i) \in [0,m]$ by setting $m$ thresholds. For $m=2$, class-0 is assigned to pixel $i$ if $a(x_i)<\theta_1$, class-1 if $\theta_1 \leq a(x_i)<\theta_2$, and class-2 if $\theta_2 \leq a(x_i)$, see Fig.~\ref{fig:act-2nested}(b). The optimal values for $\theta_{\{1,2\}}$ can either be determined through validation or preset, e.g. to 0.5 and 1.5. 

\subsection{Loss functions\label{sec:losses}}
Standard cross-entropy takes probability maps as input, 
and therefore cannot be used directly after multi-level activation, as $a(x)\in[0,m]$.
To this end, we introduce different loss functions to accommodate this new activation, that are inspired both from regression and standard multi-class classification.

\subsubsection{Sum of Squared Error loss.}
Considering the segmentation of nested classes as a regression problem, where the output map should be as close as possible to a layered-cake structure, we first propose the following Sum of Squared Error~(SSE) loss
\begin{equation}
\mathcal{L}_{\rm{SSE}} = -\frac{1}{N_{\textrm{tot}}} \sum_{\textrm{pixels}\,  i} \left[ a(x_i) - c_i \right]^2 \, ,
\label{eq:loss-SSE}
\end{equation}
where $N_{\textrm{tot}}$ is the total number of pixels, $x_i$ the CNN output for pixel $i$ and $c_i \in \{0,1,...,m\}$ the corresponding target label, chosen consistently with Eq.~\eqref{eq:act-2nested}.

\begin{figure}[tbp]
\begin{overpic}[width=0.45\textwidth]{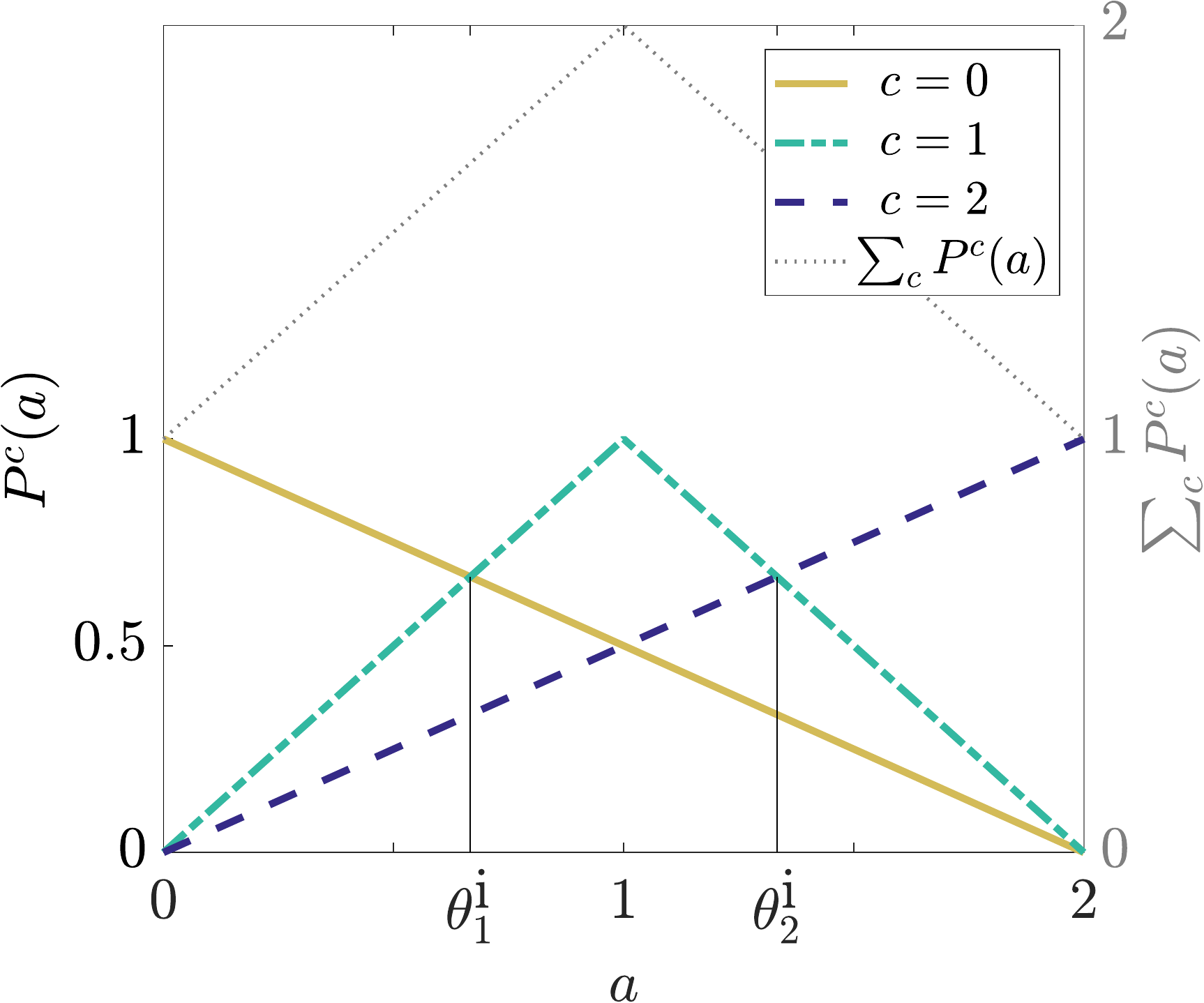}
 \put (0,78) {a)} \end{overpic}
\hfill
\begin{overpic}[width=0.45\textwidth]{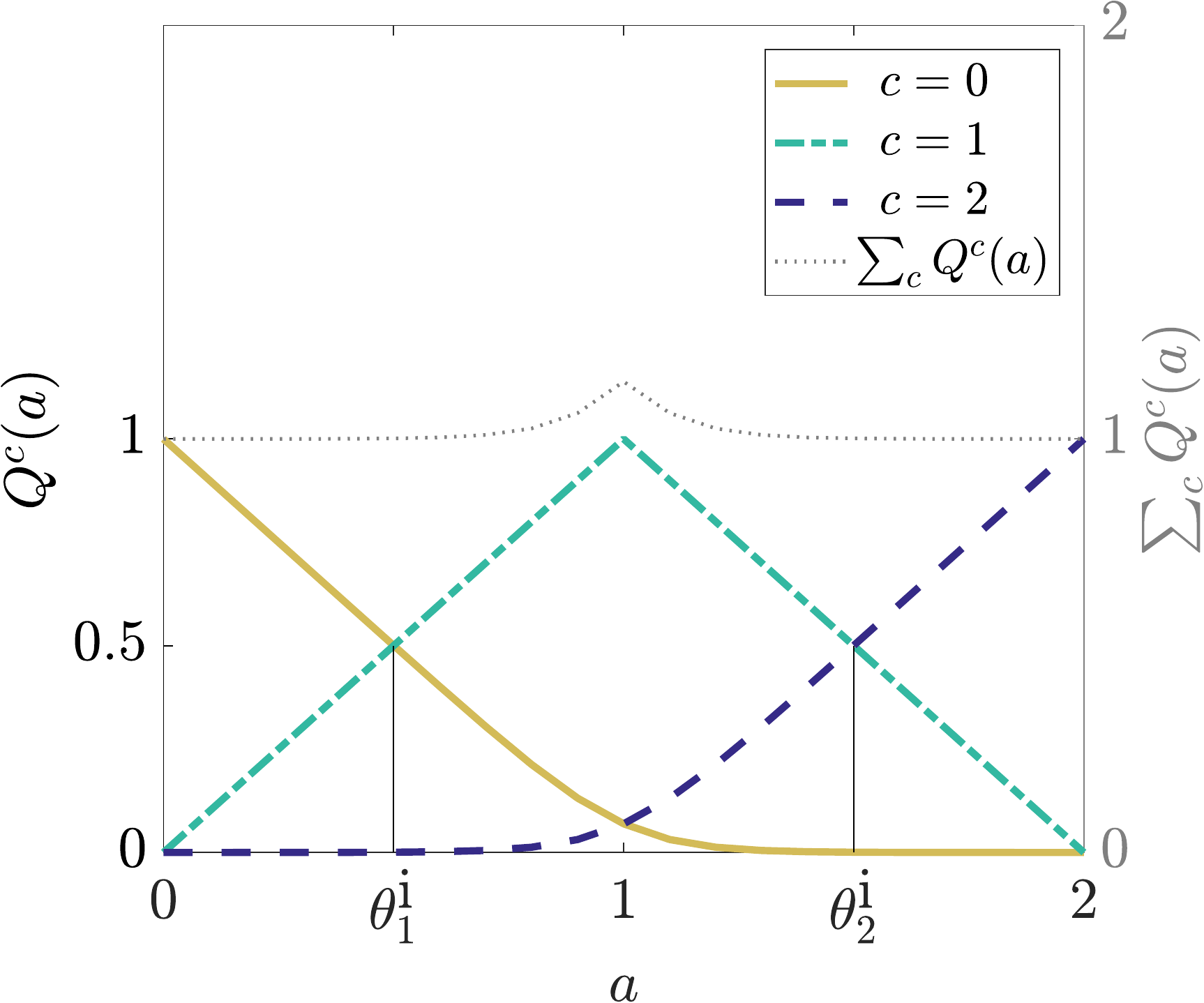}
 \put (0,78) {b)} \end{overpic}
\vspace{-0.2cm}
\caption{\emph{Class-`probabilities'.} Functions to map the output of the activation layer $a$ to pseudo-probabilities. (a) $P^{c}(a)$ for the MCE loss, see Eq.~\eqref{eq:transfo-MCE}, and (b) $Q^{c}(a)$ for the NCE loss, with $t=10$, see Eq.~\eqref{eq:transfo-softplus}.}
\label{fig:prob-2nested}
\end{figure}

\subsubsection{Modified and Normalized Cross-Entropy losses.}
Considering the problem from a multi-class classification perspective, we combine the multi-level activation with cross-entropy loss. We therefore need to map the activation $a(x) \in [0,m]$ to the interval $[0,1]$ to mimic class-probabilities. For each class $c$, this mapping should peak at the target value $c$, which becomes the attractor during training.
Focussing on the two-level case, we first propose the mapping
\begin{align}
P^{c=0}(a) & = 1-a/2 \, , \nonumber\\
P^{c=1}(a) & = 1-|1-a| \, , \label{eq:transfo-MCE} \\
P^{c=2}(a) & = a/2 \, . \nonumber
\end{align}
Those pseudo-class-probabilities are illustrated in Fig.~\ref{fig:prob-2nested}(a), 
but note that they are not strict probabilities, as they do not respect the addition rule ($\sum_c P^{c}(a) = 2-|1-a| \neq 1$).
With this transformation, the activation map can be integrated into a Modified Cross-Entropy~(MCE) loss
\begin{equation}
\mathcal{L}_{\rm{MCE}} = -\frac{1}{N_{\textrm{tot}}} \sum_{\textrm{pixels}\,  i} \sum_{\textrm{classes}\, c}\omega^c y^c_i \log \left( P^c[a(x_i)] \right) \, ,
\label{eq:loss-MCE}
\end{equation}
where $y^{c'}_i=1$ for the ground-truth label $c'$ of pixel $i$ and $y^{c \neq c'}_i=0$ otherwise.
The $P^{c}(a)$ functions display different slopes, see Fig.~\ref{fig:prob-2nested}(a), which biases the training process towards class-1, that has a higher slope and will be favored in backpropagation.
To compensate, we add the class-weights $\omega^c$ in Eq.~\eqref{eq:loss-MCE}, that are chosen to be proportional to the inverse of the number of pixels of each class in the training set, $\omega^c = N_{\textrm{tot}}/N_{c}$. 

To make up for the drawbacks of MCE, we propose a second transformation
\begin{align}
Q^{c=0}(a) & = \textrm{s}(1-a) \, , \nonumber\\
Q^{c=2}(a) & = \textrm{s}(a-1) \, ,  \label{eq:transfo-softplus}
\end{align}
where $Q^{c=1}(a)  = 1-|1-a|$ is unchanged, and we use the softplus function $\textrm{s}(x)=\frac{1}{t} \log \left( 1+e^{tx} \right)$, a smoothed version of the rectifier.
Unlike in MCE, where the slopes were biased towards class-1, the $Q^{c}(a)$ functions, which are shown in Fig.~\ref{fig:prob-2nested}(b) for $t=10$, have the same slope around the maximum.
Furthermore, they are asymptotically normalized, as we have $\sum_c Q^{c}(a) \to_{t \to \infty} 1$.
From there we define the Normalized Cross-Entropy~(NCE) loss
\begin{equation}
\mathcal{L}_{\rm{NCE}} = -\frac{1}{N_{\textrm{tot}}} \sum_{\textrm{pixels}\,  i} \sum_{\textrm{classes}\, c} y^c_i \log \left( Q^c[a(x_i)] \right) \, ,
\label{eq:loss-NCE}
\end{equation}
which leads to a balanced training.

The generalization to a higher number of nested classes is possible, and is presented explicitely for four classes in the Supplementary Material.
We also present how MCE and NCE can be combined with standard cross-entropy, by the introduction of more output channels.
For this reason, the use of those cross-entropy based losses, albeit counter-intuitive in the absence of soft-max activation and seemingly more convoluted than SSE, is of great interest:
it enables the encoding of any hierarchical tree of topologically-nested and mutually-exclusive classes in a CNN.

\section{Detection of nuclei in cells\label{sec:appli-nuclei}}
\subsubsection{Dataset and training strategy.}
\begin{figure}[tb]

\belowbaseline[5pt]{\begin{overpic}[width=4cm]{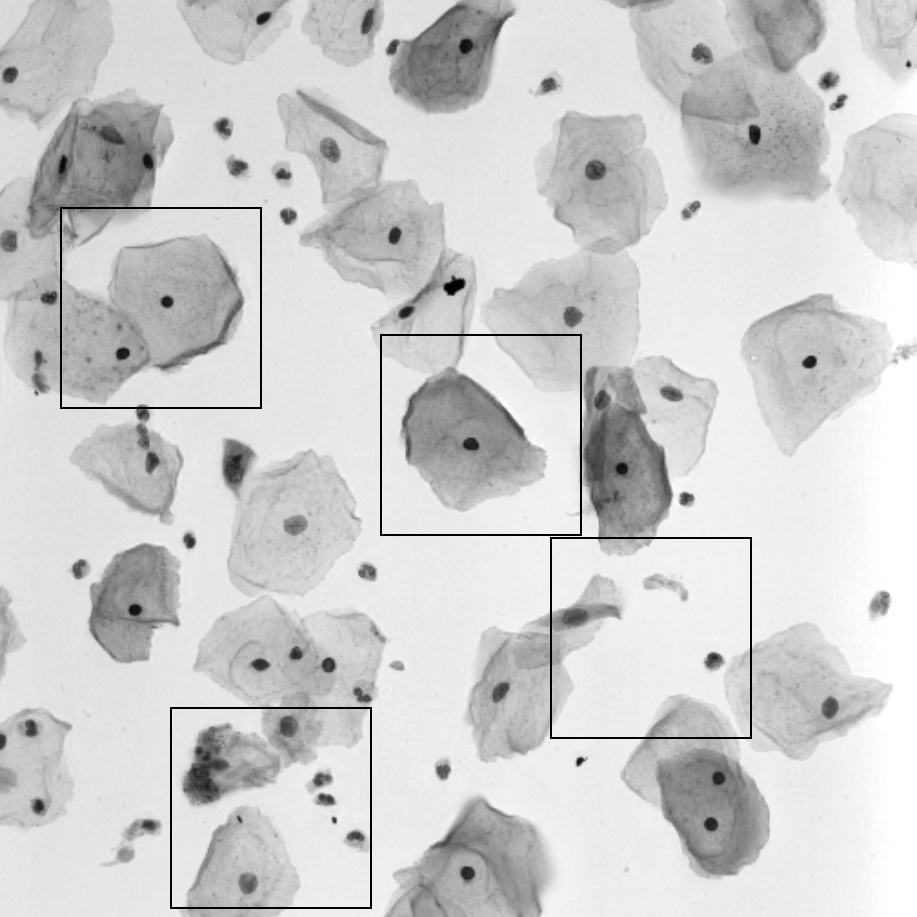}
 \put (2,92) {a)}
 \put (7,71) {3)} 
 \put (42,57) {1)}
 \put (60.5,35) {2)}
 \put (19,16) {4)}\end{overpic}} 
\hfill \belowbaseline[0pt]{\begin{overpic}[width=1.8cm]{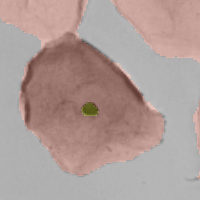}
 \put (2,84) {b1)} 
 \put (-20,3) {\rotatebox{90}{\emph{ground truth}}} \end{overpic}~
\begin{overpic}[width=1.8cm]{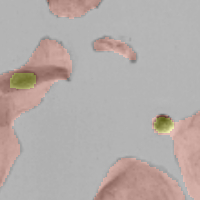}
 \put (2,84) {b2)} \end{overpic}~
 \begin{overpic}[width=1.8cm]{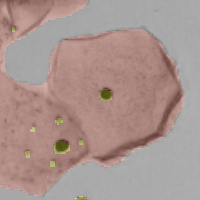}
 \put (2,84) {b3)} \end{overpic}~
\begin{overpic}[width=1.8cm]{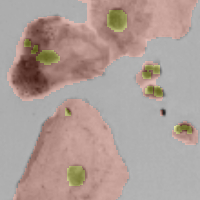}
 \put (2,84) {b4)} \end{overpic}}
 
\vspace{-2.37cm}
\vspace{0.05cm}
 
\vspace{0.058cm}
\hspace{4.565cm}
\begin{overpic}[width=1.8cm]{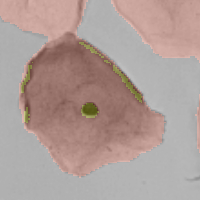}
 \put (2,84) {c1)}
 \put (-15,7) {\rotatebox{90}{7.5$\times$$10^3$ it.}} 
 \put (17.5,48) {\rotatebox{180}{\MVRightarrow}} 
 \put (17.5,60) {\rotatebox{155}{\MVRightarrow}} 
 \put (60,80) {\rotatebox{-145}{\MVRightarrow}} \end{overpic} 
\begin{overpic}[width=1.8cm]{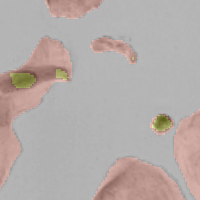}
 \put (2,84) {c2)} 
 \put (40,67.5) {\rotatebox{-180}{\MVRightarrow}}  \end{overpic} 
\begin{overpic}[width=1.8cm]{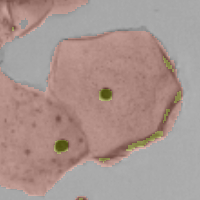}
 \put (2,84) {c3)}
 \put (60,40) {\rotatebox{-40}{\MVRightarrow}} 
 \put (70,55) {\rotatebox{20}{\MVRightarrow}} 
 \put (47.5,35) {\rotatebox{-90}{\MVRightarrow}} \end{overpic} 
\begin{overpic}[width=1.8cm]{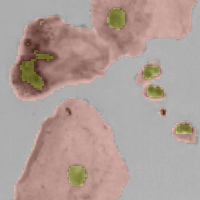}
 \put (2,84) {c4)} \end{overpic} 
 
\vspace{0.04cm}
\vspace{-0.02cm}

\hspace{4.565cm}
\begin{overpic}[width=1.8cm]{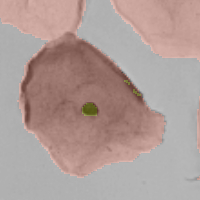}
 \put (2,84) {d1)}
 \put (-30,65) {\rotatebox{90}{\emph{multi-class}}}
 \put (-15,7) {\rotatebox{90}{30$\times$$10^3$ it.}}
 \put (65,68) {\rotatebox{-145}{\MVRightarrow}}  \end{overpic} 
\begin{overpic}[width=1.8cm]{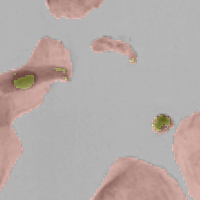}
 \put (2,84) {d2)} 
 \put (40,67.5) {\rotatebox{-180}{\MVRightarrow}}  \end{overpic} 
\begin{overpic}[width=1.8cm]{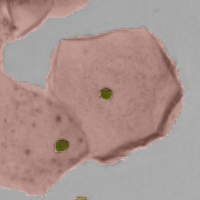}
 \put (2,84) {d3)}
 \put (28,28) {\circle{36}} \end{overpic} 
\begin{overpic}[width=1.8cm]{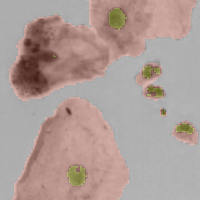}
 \put (2,84) {d4)}
 \put (20,72) {\circle{18}} \end{overpic} 

\vspace{0.04cm}
\vspace{0.05cm}

\hspace{4.565cm}
\begin{overpic}[width=1.8cm]{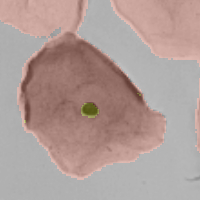}
 \put (2,84) {e1)}
 \put (-15,7) {\rotatebox{90}{7.5$\times$$10^3$ it.}} \end{overpic} 
\begin{overpic}[width=1.8cm]{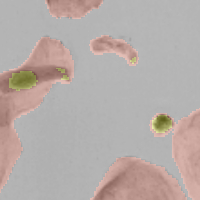}
 \put (2,84) {e2)}
 \put (40,67.5) {\rotatebox{-180}{\MVRightarrow}}   \end{overpic} 
\begin{overpic}[width=1.8cm]{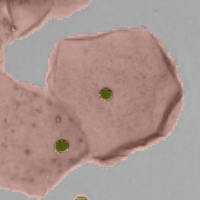}
 \put (2,84) {e3)} \end{overpic} 
\begin{overpic}[width=1.8cm]{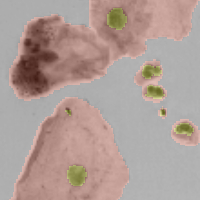}
 \put (2,84) {e4)}
 \put (20,72) {\circle{18}} \end{overpic} 
 
\vspace{0.04cm}
\vspace{-0.02cm}

\hspace{4.565cm}
\begin{overpic}[width=1.8cm]{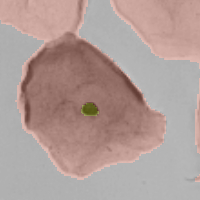}
 \put (2,84) {f1)}
 \put (-30,55) {\rotatebox{90}{\emph{our approach}}}
 \put (-15,7) {\rotatebox{90}{30$\times$$10^3$ it.}} \end{overpic} 
\begin{overpic}[width=1.8cm]{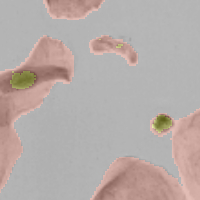}
 \put (2,84) {f2)} \end{overpic} 
\begin{overpic}[width=1.8cm]{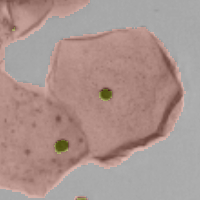}
 \put (2,84) {f3)}
 \put (28,28) {\circle{36}} \end{overpic} 
\begin{overpic}[width=1.8cm]{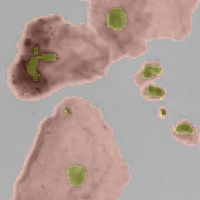}
 \put (2,84) {f4)}\end{overpic} 

\vspace{-0.2cm}
 \caption{\emph{Examples of nuclei segmentation}. (a) Selected regions of a bright-field microscopy image are overlaid with (b) the ground truth segmentation, (c-d) results from standard multi-class segmentation after 7.5 and 30$\times10^3$ iterations respectively, and (e-f) results from multi-level activation with MCE loss, our best performing method, after 7.5 and 30$\times10^3$ iterations respectively.
 The cell- (respectively nuclei-) class is displayed in red (resp. green). 
 Arrows indicate false positives and circles false negatives.
 }
\label{fig:res-nuclei}
\end{figure}
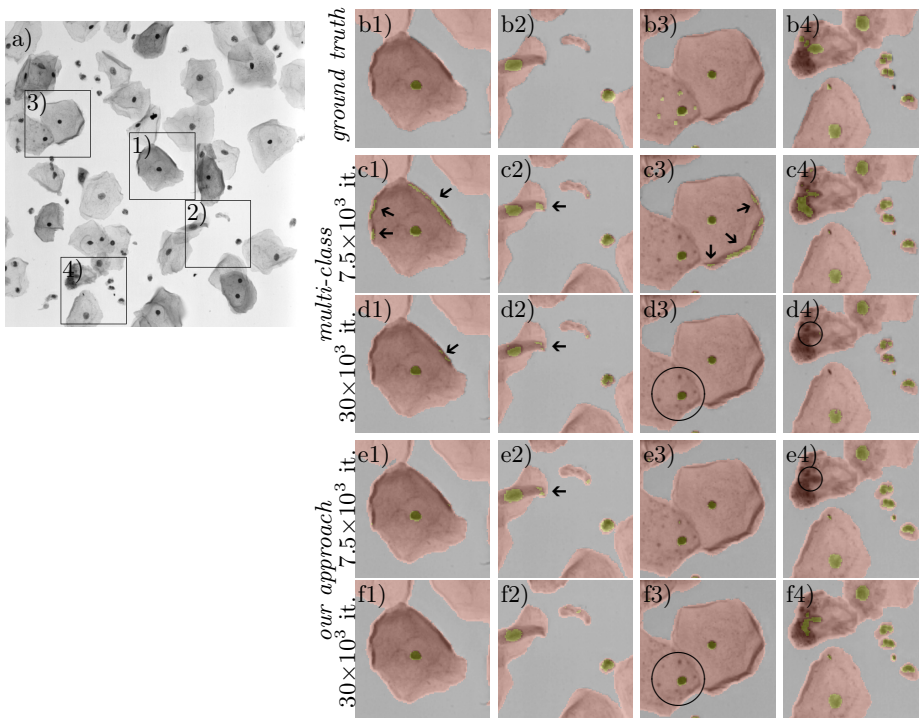

We benchmark our method on the 2018 Data Science Bowl competition from Kaggle, 
whose challenge is to detect nuclei in cell images from different microscopy modalities.
The simultaneous segmentation of cells and nuclei is an application of nested classes, as we have nuclei (class-2) $\subset$ cells (class-1) $\subset$ background (class-0).
We select all bright-field microscopy images, on which the cells are clearly visible, and complement the available nuclei segmentation by a manual segmentation of the cell bodies (see Fig.~\ref{fig:res-nuclei}).
Another challenge from this competition is to deal with the limited number of training images in this modality ($N_{\textrm{im}}=16$).
We therefore rely on online data augmentation, and perform random flips, warping, rotations, translations and rescaling of the images at each training epoch.
The images, resized to $512 \times 512$ pixels, are then fed into a U-Net like architecture~\cite{Ronneberger2015}.

We perform 4-fold cross-validation and a rotating testing scheme:
we split the dataset into 4 subsets to perform cross-validation and further split each validation set into 4, for testing, such that we always train on 12 images, validate on 3 and test on one, in a 16-fold rotating fashion. 
For each fold, we select the threshold value $\theta_2$ giving the best Dice score on the validation set and compute the scores of the test image after 30$\times10^3$ iterations. 
Note that due to online data augmentation, the model does not tend to overfit.

\leavevmode
\subsubsection{Results.}
Qualitative results for standard multi-class segmentation and multi-level activation with MCE loss, after 7.5$\times10^3$ iterations and after convergence, are presented in Fig.~\ref{fig:res-nuclei}.
Early on in the training process, multi-class segmentation outputs false positives for the nuclei class, in darker regions of the cells, even very close to the cell border, see Figs.~\ref{fig:res-nuclei}(c1)-(c3).
Those slowly disappear in the training process, as the CNN learns the relationship between classes, although some imperfections remain, see Fig.~\ref{fig:res-nuclei}(d1)-(d2).  
This is strongly disfavored by multi-level activation, as the output map $\{x_i\}$ would have to oscillate on very short length scales, and we see that most such artefacts do not appear on Figs.~\ref{fig:res-nuclei}(e)-(f).
On the contrary, the spatial regularisation provided by multi-level activation can lead to false negatives at the beginning of the training process, as on Fig.~\ref{fig:res-nuclei}(e4), which are then detected by the converged model, see Fig.~\ref{fig:res-nuclei}(f4). 
The circles on Figs.~\ref{fig:res-nuclei}(d3) and (f3) outline difficult cases which are missed by both methods.

\begin{figure}[tb]
\vspace{-0.75cm}
\belowbaseline[-90pt]{\includegraphics[width=0.42\textwidth]{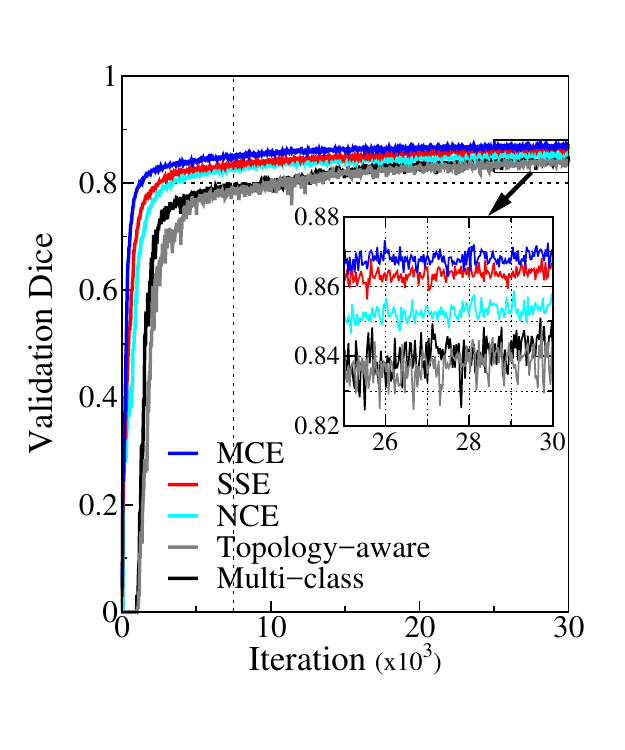}}
\hfill
\begin{tabular}{| c | c | c |}
\hline
 {\bf Model} & $\bold{\theta_2}$ & {\bf Test Dice} \\  \hline
\hline
\textsc{i} - Multi-class  & -- & 0.839 (0.055) \\  \hline
\textsc{ii} - Topology-aware~\cite{BenTaieb2016} & -- & 0.842 (0.055) \\ \hline
\hline
\textsc{iii} - NCE, Eq.~\eqref{eq:loss-NCE} & $\theta_2^{\textrm{i}}= 1.5$ & 0.841 (0.058) \\ \hline
\textsc{iv} - SSE, Eq.~\eqref{eq:loss-SSE} & $\theta_2^{\textrm{i}}=1.5$ & \cellcolor{mygray!50}0.863 (0.051)\\ \hline
\textsc{v} - MCE, Eq.~\eqref{eq:loss-MCE} & $\theta_2^{\textrm{i}}=4/3$ & 0.844 (0.049)\\ \hline
\hline
\textsc{vi} - NCE, Eq.~\eqref{eq:loss-NCE} & 1.31 (0.09) & \cellcolor{mygray!50}0.853 (0.056) \\ \hline
\textsc{vii} - SSE, Eq.~\eqref{eq:loss-SSE} & 1.39 (0.08) &\cellcolor{mygray!50}0.863 (0.053) \\ \hline
\textsc{viii} - MCE, Eq.~\eqref{eq:loss-MCE} & 1.74 (0.05) & \cellcolor{mygray!50}0.868 (0.051)\\
\hline
\end{tabular}
\vspace{-0.75cm}
\captionsetup{labelformat=andtable}
\caption{\emph{Scores for nuclei segmentation}.
{\bf Fig.:} Mean validation Dice scores for the methods listed in the table (rows \textsc{i}, \textsc{ii} and \textsc{vi} to \textsc{viii}).
{\bf Table:} Mean test Dice scores for standard multi-class (row \textsc{i}), the `topology-aware' loss~\cite{BenTaieb2016} (row \textsc{ii}) and multi-level activation (rows \textsc{iii}-\textsc{viii}). 
In rows  \textsc{iii} to \textsc{v} we use the threshold $\theta_2^{\textrm{i}}$. In rows \textsc{vi} to \textsc{viii}, $\theta_2$ is selected during validation and we report its mean value. Grey cells highlight significant improvement over multi-class, i.e. p-values~$<0.05$ using the paired samples Wilcoxon test.}
\label{fig:res-kaggle}
\captionlistentry[table]{}
{\makeatletter\edef\@currentHref{table.caption.\the\c@table}\label{tab:res-kaggle}}
\end{figure}

We now compare quantitatively two soft-max based methods --~standard multi-class segmentation and the `topology-aware' method from Ref.~\cite{BenTaieb2016}~-- with our multi-level activation layer, combined with the three losses introduced in Sec.~\ref{sec:losses}.
The validation Dice scores for the nuclei class are shown in Fig.~\ref{fig:res-kaggle} for each method of this benchmark.
The corresponding test scores are reported in Table~\ref{tab:res-kaggle}, for (i)~a pre-determined value $\theta_2^{\textrm{i}}$ and (ii)~the value of $\theta_2$ selected during validation.
In (i), we choose the \emph{a priori} inferred value $\theta_2^{\textrm{i}}=1.5$ for NCE and SSE, and $\theta_2^{\textrm{i}}=4/3$ for MCE, the values where $P^1$ and $P^2$, respectively $Q^1$ and $Q^2$, intersect (see Fig.~\ref{fig:prob-2nested}).
While all methods perform comparatively well for cell segmentation, with mean test Dice scores ranging from 0.977\,(0.01) to 0.979\,(0.01), there are quantitative discrepancies in the nuclei segmentation.
The two soft-max based methods perform on par: we report a mean test Dice score of 0.839\,(0.055) for standard multi-class, which we consider as our baseline in the following, and of 0.842\,(0.055) for the `topology-aware' loss~\cite{BenTaieb2016}.
Indeed, multi-class is not plagued by the detection of nuclei outside cells, but rather outputs false positives near the border of the cell, as we have seen above, a problem which is not addressed by the method of Ref.~\cite{BenTaieb2016}.

In Fig.~\ref{fig:res-kaggle}, we see that our three proposed methods converge much faster, and outperform the soft-max based ones during validation.
Our methods indeed cross the 0.8 validation Dice score after 1.3 to 3.2$\times 10^3$ iterations, whereas the soft-max-based methods need more than three times as much iterations to achieve this accuracy.
Using $\theta_2^{\textrm{i}}$, we improve the test Dice score by 0.2 and 0.5 for $\mathcal{L}_{\rm{NCE}}$ and $\mathcal{L}_{\rm{MCE}}$ and by  2.4 points with $\mathcal{L}_{\rm{SSE}}$.
The paired samples Wilcoxon test gives a p-value of 0.0008 for $\mathcal{L}_{\rm{SSE}}$ vs multi-class, confirming the significance of this improvement.
Threshold selection during validation improves the $\mathcal{L}_{\rm{NCE}}$ and $\mathcal{L}_{\rm{MCE}}$ results further, exceeding the multi-class score by 1.4 and 2.9 Dice points respectively (with p-values~$0.007$ and $0.001$).
$\mathcal{L}_{\rm{MCE}}$ therefore gives the best model of this benchmark.
Note that using $\mathcal{L}_{\rm{MCE}}$ without reweighting [i.e. $\omega^c=1$ in Eq.~\eqref{eq:loss-MCE}] strongly undersegments nuclei, confirming that the training is then biased towards class-1, as anticipated in Sec.~\ref{sec:losses}.
Furthermore, with $\omega^c\propto1/N_{\textrm{c}}$, the thresholds selected by validation, with mean $\theta_2^{\textrm{mean}}=1.74$, significantly differ from $\theta_2^{\textrm{i}}=4/3$, indicating that class-2 might now be overfavored. 
We retrospectively verified that all other models do not benefit from the application of the weighting scheme $\omega^c \propto 1/N_{c}$.

\subsubsection{Discussion.}
Our proposed multi-level activation layer, greatly speeds up learning and outperforms the soft-max based methods. 
It indeed permits to significantly improve the nuclei test Dice scores in all cases (with p-values~$<0.007$). 
This novel activation layer introduces thresholds in the multi-class classification context, which can be adjusted at validation time, leading to a significant performance gain for $\mathcal{L}_{\rm{NCE}}$ and $\mathcal{L}_{\rm{MCE}}$.
But this is not the only benefit of our method, as all proposed loss functions 
outperform multi-class without threshold adjustment, significantly for $\mathcal{L}_{\rm{SSE}}$, proving that our regression-like method is better suited to the problem.

\section{Conclusion}
In this work, we have proposed a new paradigm for multi-class segmentation with topological constraints of inclusion. It consists in a novel multi-level activation layer and three matching loss functions, based on regression and cross-entropy loss.
This scheme can be implemented in any network architecture, with minimal changes.
We benchmarked our method on the segmentation of nuclei in bright-field microscopy images, giving significant improvement and speed-up over the soft-max based methods.
In the Supplementary Material, we provide a second benchmark on liver lesions segmentation, in a larger and more imbalanced dataset, with the same conclusions.
We expect those results to transfer to other tasks with nested classes, as nothing was handcrafted for the problems at stake.
$\mathcal{L}_{\rm{MCE}}$ turned out to be the best performing loss in this paper, but the other losses also perform well, and might be better suited for different applications.

Informing the network on the relations between classes with the multi-level activation thus permits to train on less data, which is often crucial in biomedical applications.
Finally, as shown in the Supplementary Material, the multi-level activation layer and the associated losses can be straightforwardly generalized to a deeper nesting hierarchy.
We also show how $\mathcal{L}_{\rm{MCE}}$ and $\mathcal{L}_{\rm{NCE}}$ can be used alongside normal cross-entropy to segment nested classes together with further classes without topological prior. This enables the encoding of any tree of prior relations of containment between classes.
\bibliographystyle{splncs}

\end{document}